\documentclass{article}
\usepackage{spconf,amsmath,graphicx}
\usepackage{amssymb,amsfonts,bm,mathtools}
\usepackage{cite}
\usepackage{xcolor,colortbl}
\usepackage{booktabs,multirow}
\usepackage{balance}
\usepackage[hidelinks]{hyperref}

\definecolor{offwhite}{RGB}{244,243,238}
\newcommand{\sh}{\cellcolor{offwhite}}

\title{D-MOD: DIFFUSION MODE ESTIMATION FOR ACCELERATED MRI}

\name{Md Kawsar Ahmed$^{1,2}$ and Mathews Jacob$^{1}$}
\address{$^{1}$University of Virginia, Charlottesville, VA, USA/\\
  $^{2}$Bangladesh University of Engineering and Technology, Dhaka, Bangladesh.\\
  \{ngd6kg, mjacob\}@virginia.edu}

\begin{document}
\maketitle

\begin{abstract}
Diffusion models are powerful priors for accelerated MRI reconstruction, but
most current methods are posterior samplers, which returns a random draw from the posterior. A reproducible and high-quality reconstruction often requires averaging many samples, resulting in a high
computational cost. We introduce \textbf{D-MOD}, a fast algorithm that instead
provides a posterior \emph{mode}. It uses a minorize--maximize algorithm on a lower bound of the smoothed log-posterior, which provably increases
the objective. Each iteration draws $S$ noisy samples
of the current estimate at the current noise level, denoises them with a
pretrained diffusion model, and combines their average with the measured
k-space through a conjugate-gradient (CG) solve. We anneal the noise levels, as the distance between the bound and its target decays quadratically with noise. On fastMRI brain (T1, T2, FLAIR), a single D-MOD
reconstruction exceeds, in PSNR, the $10$-sample finite-MMSE estimates of DPS,
DAPS and SNORE-P and stays within $0.31$\,dB of the strongest
baselines (DiffPIR, DDS). This approach thus provides a fast deterministic reconstruction algorithm ($11$--$73\times$ faster than MMSE estimates), which is useful in a clinical setting.
\end{abstract}


% =====================================================================
\section{Introduction}
\label{sec:intro}

Diffusion models are powerful priors for accelerated MRI.
Trained once on fully sampled data, they generalize across contrasts and
anatomies without task-specific retraining~\cite{ho2020ddpm,song2021score,%
karras2022edm}. Since their first application to
MRI~\cite{Jalal2021scoreALD,chung2022scoreMRI}, the dominant approach has been
to inject data consistency corrections along the reverse trajectory of the diffusion models. As the exact correction is intractable, the methods
mainly differ on the approximations to compute it \cite{Chung2023DPS,Zhang2025DAPS,zhu2023diffpir,chung2023dds}. All of these methods return a \emph{sample} from the posterior distribution, depending on the random initialization from the prior or the intermediate noise realizations. The reconstructions vary significantly with the random seed, and a reproducible result requires averaging several
samples to approximate the MMSE estimate. This averaging compounds an already large
iteration count of diffusion posterior sampling, and may blur fine detail. In most clinical settings, the acceleration is set to be modest to minimize unacceptable hallucinations, when the posterior is unimodal. In these settings, a fast and deterministic solution that is the most plausible one is preferable to posterior sampling. 
%In these settings, a mode estimate is the most plausible point, and a single fast, reproducible reconstruction is preferable to a set of samples.

An alternative is to estimate a mode of the posterior directly. Plug-and-play
(PnP) MAP methods~\cite{venkatakrishnan2013pnp,kamilov2022pnp,laumont2022pnpmap,%
romano2017red} embed a learned denoiser as an implicit prior and return a
deterministic estimate with noise annealing~\cite{renaud2024snore,chand2024multi}. Recently, renoise--denoise regularization~\cite{renaud2024snore,renaud2025proxconv} was introduced to further reduce the impact of
distribution mismatch. We note some limitations of these methods that restrict their efficiency. First, the
ascent step size is bounded by the Lipschitz constant of the denoiser residual, which is not available in closed form. Empirical Lipschitz bounds are conservative and
translate to small steps, while constraining the Lipshitz constant of the denoiser during
training~\cite{ryu2019pnptrained} degrades the denoising
performance~\cite{hurault2022gradstep}. Second, the prior expectation is approximated by Monte Carlo in \cite{renaud2024snore,renaud2025proxconv}; the
step-sizes must be small enough to average that noise \emph{across}
iterations, leaving the final estimate seed-dependent.

We introduce D-MOD, an annealed mode-estimation algorithm that removes both obstacles. Each
iteration adds $S$ independent noise realizations at the current noise level to the
iterate, denoises each of them and averages the results, and then combines that average
with the measured k-space by a CG least-squares solve. Together these steps form a minorize--maximize (MM) algorithm to maximize a lower bound of the smoothed
log-posterior, which is guaranteed to increase the lower-bound monotonically; the denoised average is the gradient of the prior term and the CG solve returns the exact maximizer of a quadratic minorizer. Our analysis shows that the lower-bound closely approximates the log-posterior at the modes at each noise scale. Additionally, it converges to the true posterior as the noise level approaches zero. The averaging of the denoised images at each iteration results in a $O(1/S)$ reduction in the prior variance. The algorithm thus offers a mode estimate with low variance, which can serve as a fast and more deterministic alternative to diffusion posterior sampling methods in clinical applications. 

% =====================================================================
\section{Method}
\label{sec:method}

%\subsection{Problem and the Smoothed Prior}
MRI measurements are modeled as $y=Ax+n$, $n\sim\mathcal{N}(0,\beta_y^2I)$,
with $A\in\mathbb{C}^{m\times p}$ the undersampled multicoil Fourier operator,
$x\in\mathbb{C}^p$ the image and $\beta_y^2$ the noise variance. Bayesian methods consider  $\log p(x\mid y) = J(x)$:
\begin{equation}
  J(x)=\log p(y\mid x)+\log p(x).
  \label{eq:posterior}
\end{equation}
The measurement term $\log p(y\mid x)=-\|Ax-y\|_2^2/(2\beta_y^2)+c$ is closed
form and concave. The true prior is often non-concave and often supported on a
set of measure zero (manifold hypothesis), and hence $\log p=-\infty$ almost everywhere; this affects the convergence of all gradient-based algorithms as the prior gradient will be zero, when initialized away from the data support. Diffusion models overcome the above gradient issues by using Gaussian-smoothed priors
$p_\sigma=p*\mathcal{N}(0,\sigma^2I)$.
With $\hat{x}_0(u):=\mathbb{E}[x\mid x+\sigma\varepsilon=u]$ the MMSE denoiser
at level $\sigma$, Tweedie's identity relates it to the score,
\begin{equation}
  \hat{x}_0(u)=u+\sigma^2\nabla\log p_\sigma(u),
  \label{eq:tweedie}
\end{equation}
and a trained diffusion model supplies an estimate of $\nabla\log p_\sigma$ for
the whole family $\{p_\sigma\}$. 

Two limitations restrict the direct use of diffusion models for mode estimation. \emph{(L1)} Diffusion models supply the score estimates, but not the log density values. While the score estimates can enable gradient-based algorithms, it is challenging to perform objective comparisons, line
searches, and choose step-sizes that guarantee convergence. \emph{(L2)} Score estimates are reliable only at noisy inputs:
trained on samples of $p_{\sigma_t}$, which concentrate at distance
$\approx\sigma_t\sqrt{d}$ from the clean data. In particular, the score estimates are not well defined at samples $x\sim p(x)$ itself, which restrict the direct utility of diffusion models in this context.

%To overcome the above challenges, we propose to  replace $J$ at each noise level by a surrogate that is maximized before $\sigma_t$ is lowered. Three properties make one usable: \textbf{(P1)} its gradient is expressible through score evaluations alone, at noisy inputs (from L1, L2); \textbf{(P2)} it lies below the smoothed log-posterior with a gap quantified at finite $\sigma_t$ that vanishes as $\sigma_t\to0$; and \textbf{(P3)} its curvature admits a known lower bound $\nabla^2\succeq-cI$ with $c$ in closed form, so a quadratic minorizer exists everywhere and can be maximized without a step size. (P3) separates this formulation from gradient-based schemes, which need an \emph{upper} curvature bound --- aLipschitz constant a learned denoiser does not provide in closed form.
% ---------------------------------------------------------------- ALGORITHM
\begin{figure}[t]
\footnotesize
\hrule\vspace{2pt}
\noindent\textbf{Algorithm 1} D-MOD
\vspace{2pt}\hrule\vspace{3pt}
\noindent\textbf{Input:} annealing steps $T$, samples $S$, inner MM iterations
$K$, CG iterations $L$, data $y$, operator $A$, noise variance $\beta_y^2$,
schedule $\sigma_{t_0}>\cdots>\sigma_{t_{T-1}}$.\\[1pt]
\textbf{Initialize:} $x^{(0)}=0$.
\vspace{3pt}\hrule\vspace{3pt}
\noindent\textbf{for} $n=0$ \textbf{to} $T-1$\\
\hspace*{1em}$\sigma_t\leftarrow\sigma_{t_n}$,\quad $x_0^{(n)}\leftarrow x^{(n)}$\\
\hspace*{1em}\textbf{for} $k=0$ \textbf{to} $K-1$\\
\hspace*{2em}{\textbf{Renoise:}} $x_{t,s}\sim\mathcal{N}(x_k^{(n)},\sigma_t^2I)$, \; $s=1,\dots,S$\\[1pt]
\hspace*{2em}{\textbf{Denoise:}} evaluate $\hat{x}_0(x_{t,s})$ for each $s$\\[1pt]
\hspace*{2em}{\textbf{Average:}} $\tilde{x}_0\leftarrow\frac{1}{S}\sum_{s=1}^{S}\hat{x}_0(x_{t,s})$\\[2pt]
\hspace*{2em}{\textbf{Maximize:}} solve by $L$ CG steps\\[2pt]
\hspace*{2em}$\Big(\tfrac{A^{\mathsf H}\!A}{\beta_y^2}+\tfrac{I}{\sigma_t^2}\Big)x_{k+1}^{(n)}=\tfrac{A^{\mathsf H}y}{\beta_y^2}+\tfrac{\tilde{x}_0}{\sigma_t^2}$\\[2pt]
\hspace*{1em}\textbf{end for}\\
\hspace*{1em}$x^{(n+1)}\leftarrow x_K^{(n)}$\\
\textbf{end for}\\
\textbf{return} $x^{(T)}$ \hfill {\footnotesize (total NFE $=T\!\cdot\!S\!\cdot\!K$)}
\vspace{2pt}\hrule
\end{figure}

% ---------------------------------------------------------------- FIGURE
\begin{figure*}[t]
  \centering
  \includegraphics[width=0.95\textwidth]{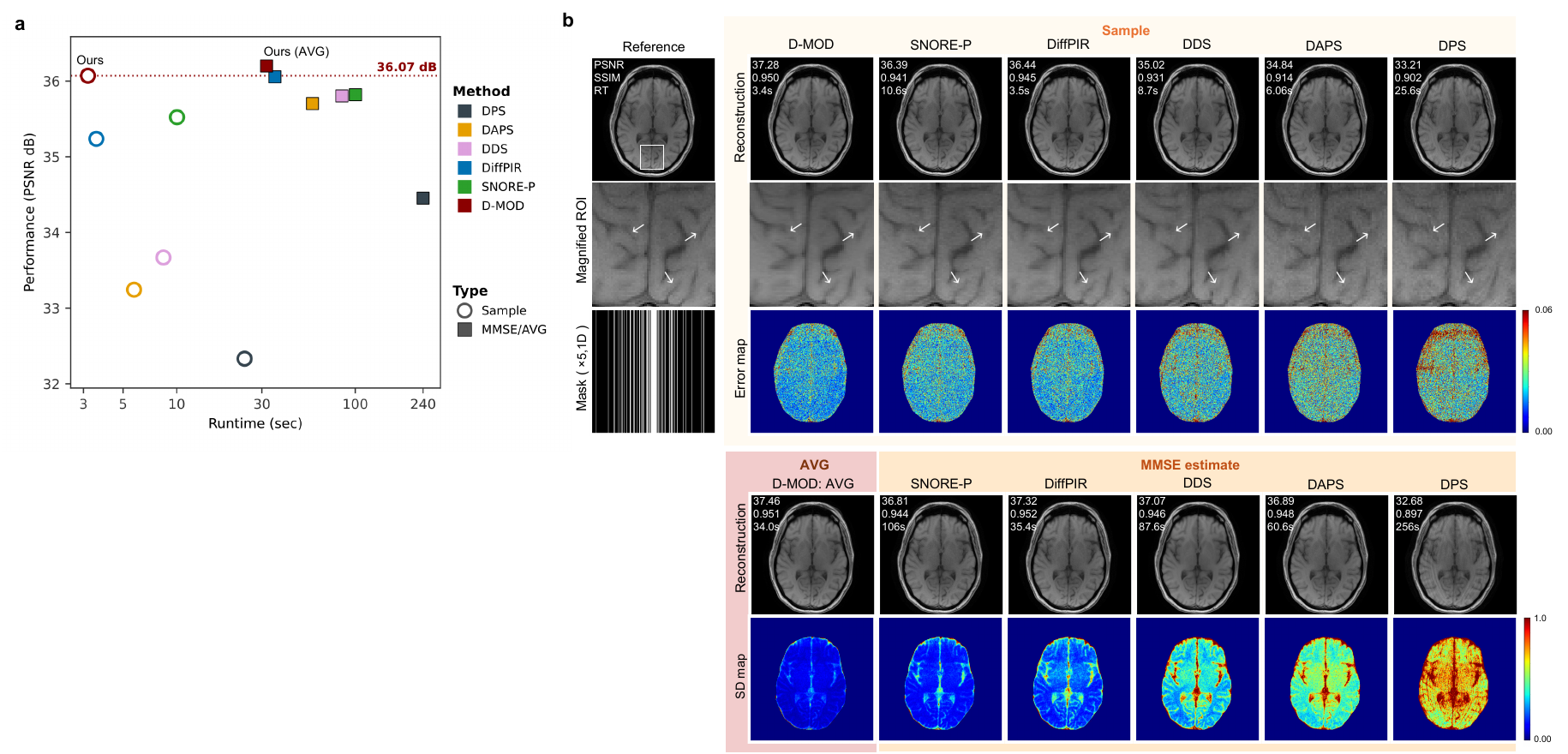}
  \caption{fastMRI brain, T1, $\times5$. \textbf{(a)} PSNR versus per-slice
runtime (log scale); circles denote single-sample reconstructions and squares
the 10-sample MMSE estimates (AVG for D-MOD); top-left is best. \textbf{(b)}
Qualitative comparison: (top) single-sample reconstructions with a magnified
ROI and absolute-error map; (bottom) 10-sample MMSE estimate with its
pixel-wise SD map (higher SD, lower reproducibility). Arrows mark artifacts or
blurred structure; PSNR (dB), SSIM and runtime (s) are shown.}
  \label{fig:main}
\end{figure*}

\subsection{A Lower Bound on the Smoothed Log-Posterior}
\label{subsec:bound}

To overcome the above challenges, we propose to  use  a lower-bound for the smoothed log posterior. 
The $\sigma_t$-smoothed prior admits the representation
$p_{\sqrt2\sigma_t}(x)=\int p_{\sigma_t}(x_t)\mathcal{N}(x;x_t,\sigma_t^2I)dx_t$,
an identity expressing that smoothing twice by $\mathcal{N}(0,\sigma_t^2I)$
equals smoothing once by $\mathcal{N}(0,2\sigma_t^2I)$. Writing it as an
expectation under an arbitrary density $q(x_t)$ and applying Jensen's
inequality yields
\begin{equation}
  \log p_{\sqrt2\sigma_t}(x)\;\ge\;
  \mathbb{E}_q\!\left[\log\frac{p_{\sigma_t}(x_t)\,\mathcal{N}(x;x_t,\sigma_t^2I)}{q(x_t)}\right].
  \label{eq:jensen}
\end{equation}
The bound is one-sided by
construction, with a KL-divergence gap of $\mathrm{KL}(q\,\|\,p_t(\cdot\mid x))$. It remains to choose $q$. We consider an isotropic kernel $q_s(x_t)=\mathcal{N}(x_t;x,s^2I)$ centered at $x$. Differentiating under the expectation leaves an average of scores at the
perturbed points $x+s\varepsilon$. We choose $s=\sigma_t$ to ensure that the the network $\nabla\log p_{\sigma_t}$ is queried at $(x+s\varepsilon) \sim p_{\sigma_t}$, for any clean $x$. Choosing
$s=\sigma_t$ gives
\begin{equation}
  G_t(x):=\mathbb{E}_{x_t\sim\mathcal{N}(x,\sigma_t^2I)}\!\left[\log p_{\sigma_t}(x_t)\right],
  \label{eq:Gdef}
\end{equation}
the stochastic denoising regularizer of Renaud \emph{et
al.}~\cite{renaud2024snore}; the derivation shows that $ G_t(x)$ is a lower bound on a smoothed log-prior. Adding the exact measurement term gives the per-level
objective
\begin{equation}
  C_t(x)=\log p(y\mid x)+G_t(x)\!\le\!
  \underbrace{\log p(y\mid x)+\log p_{\sqrt2\sigma_t}(x)}_{=:\,J_t(x)} .
  \label{eq:Ct}
\end{equation}

Three properties of $G_t$, which hold independently of the data distribution, the
architecture and the training, aid with the algorithm. First, differentiating~\eqref{eq:Gdef}
under the expectation and applying~\eqref{eq:tweedie} at every perturbed point
gives, for every $\sigma_t>0$,
\begin{equation}
  \nabla G_t(x)=\frac{\bar{x}_0(x)-x}{\sigma_t^2},\quad
  \bar{x}_0(x):=\mathbb{E}_{\varepsilon}\!\left[\hat{x}_0(x+\sigma_t\varepsilon)\right];
  \label{eq:grad}
\end{equation}
the prior gradient is obtained by perturbing the current estimate at the
current noise level, denoising, and averaging. Second, the curvature
bound~\cite{miyasawa1961,gribonval2011}
\begin{equation}
  \nabla^2G_t(x)\succeq-\sigma_t^{-2}I
  \label{eq:curv}
\end{equation}
holds at every $x$ and every $\sigma_t>0$, with a constant set by the noise
level rather than tuned. Third, the gap between the lower bound and true posterior
$\Delta_t(x):=J_t(x)-C_t(x)$ satisfies
\begin{equation}
  \Delta_t(x)=\tfrac{\sigma_t^2}{2}\big\|\nabla\log p_{\sigma_t}(x)\big\|^2+O(\sigma_t^4),
  \label{eq:gap}
\end{equation}
reducing at a stationary point $x^\star$ of $C_t$ to
$\sigma_t^2\|A^{\mathsf H}(Ax^\star-y)\|^2/(2\beta_y^4)+O(\sigma_t^4)$.
Annealing therefore drives the gap to zero at a known rate, the surrogate
agrees with the smoothed log-prior in value and gradient at the stationary
points of $\log p_{\sigma_t}$; the residual gap at convergence is small
whenever the solution agrees with the measurements. Consequently a maximizer of
$C_t$ nearly maximizes $J_t$, the shortfall being at most $\Delta_t$; if
$-\nabla^2J_t\succeq\mu I$ near the two maximizers, their distance is
$O(\sigma_t/\sqrt{\mu})$.

% ---------------------------------------------------------------- TABLE
\begin{table*}[t]
\centering
\caption{Reconstruction performance on fastMRI brain (30 subjects per contrast, 1D Cartesian).
Entries are PSNR (dB)/SSIM; for each method the shaded ``MMSE'' row is the $10$-sample finite
estimate and ``Sample'' is a single-sample reconstruction. Best value in \textbf{bold}.}
\label{tab:fastmri}
\footnotesize
\setlength{\tabcolsep}{4pt}
\renewcommand{\arraystretch}{1.0}
\begin{tabular}{@{}c c l cccccc@{}}
\toprule
\textbf{Contrast} & \textbf{Acc.} & \textbf{Type}
& \textbf{DPS} & \textbf{DAPS} & \textbf{DDS} & \textbf{DiffPIR}
& \textbf{SNORE-P} & \textbf{D-MOD$^{\dagger}$} \\
\midrule
\multirow{4}{*}{T1}
 & \multirow{2}{*}{$\times5$}
 & Sample & 32.33/0.905 & 33.24/0.915 & 33.67/0.932 & 35.24/0.947 & 35.52/0.945 & \textbf{36.07}/\textbf{0.952} \\
 & & \sh MMSE & \sh 34.45/0.939 & \sh 35.70/0.949 & \sh 35.80/0.949 & \sh 36.06/\textbf{0.954} & \sh 35.82/0.948 & \sh \textbf{36.20}/0.952 \\
 & \multirow{2}{*}{$\times10$}
 & Sample & 26.32/0.799 & 30.36/0.870 & 30.73/0.895 & 32.36/0.919 & 32.69/0.923 & \textbf{33.23}/\textbf{0.927} \\
 & & \sh MMSE & \sh 28.35/0.857 & \sh 32.78/0.922 & \sh 32.96/0.924 & \sh 33.38/\textbf{0.931} & \sh 33.14/0.928 & \sh \textbf{33.50}/0.929 \\
\addlinespace[2pt]
\multirow{4}{*}{T2}
 & \multirow{2}{*}{$\times5$}
 & Sample & 35.75/0.949 & 37.27/0.957 & 37.15/0.963 & 38.83/\textbf{0.972} & 39.07/0.971 & \textbf{39.53}/\textbf{0.972} \\
 & & \sh MMSE & \sh 36.19/0.956 & \sh 39.34/0.974 & \sh 39.00/0.970 & \sh \textbf{39.67}/\textbf{0.976} & \sh 39.39/0.972 & \sh \textbf{39.67}/0.973 \\
 & \multirow{2}{*}{$\times10$}
 & Sample & 28.77/0.855 & 33.73/0.928 & 34.32/0.942 & 35.25/0.950 & 35.02/0.950 & \textbf{35.72}/\textbf{0.951} \\
 & & \sh MMSE & \sh 29.68/0.882 & \sh 35.49/0.954 & \sh 35.86/0.951 & \sh \textbf{36.03}/\textbf{0.956} & \sh 35.41/0.954 & \sh 35.93/0.952 \\
\addlinespace[2pt]
\multirow{4}{*}{FLAIR}
 & \multirow{2}{*}{$\times5$}
 & Sample & 32.34/0.890 & 33.09/0.900 & 33.23/0.913 & 35.28/0.939 & 35.35/0.935 & \textbf{36.09}/\textbf{0.942} \\
 & & \sh MMSE & \sh 35.27/0.938 & \sh 35.75/0.941 & \sh 35.76/0.940 & \sh 36.19/\textbf{0.947} & \sh 35.72/0.939 & \sh \textbf{36.25}/0.943 \\
 & \multirow{2}{*}{$\times10$}
 & Sample & 26.33/0.772 & 29.94/0.844 & 30.25/0.867 & 32.48/0.907 & 32.63/0.908 & \textbf{33.50}/\textbf{0.913} \\
 & & \sh MMSE & \sh 30.05/0.871 & \sh 32.80/0.908 & \sh 32.85/0.909 & \sh 33.63/\textbf{0.922} & \sh 33.29/0.915 & \sh \textbf{33.75}/0.915 \\
\addlinespace[2pt]
\multicolumn{2}{@{}c}{\multirow{2}{*}{\textbf{Runtime (s)}}}
 & Sample & 25.52 & 6.06 & 8.70 & 3.78 & 10.67 & \textbf{3.50} \\
\multicolumn{2}{@{}c}{}
 & \sh MMSE & \sh 255.19 & \sh 60.59 & \sh 86.93 & \sh 37.80 & \sh 106.68 & \sh \textbf{34.95} \\
\bottomrule
\multicolumn{9}{@{}l@{}}{\scriptsize $^{\dagger}$For D-MOD, MMSE is a 10-run average of mode estimates.}\\
\end{tabular}
\end{table*}

\subsection{The Minorize--Maximize Iteration}
\label{subsec:mm}
Let $x_k$ be the current iterate at level $\sigma_t$. By~\eqref{eq:curv}, $G_t$
lies above its tangent paraboloid of curvature $-\sigma_t^{-2}$ at $x_k$.
Substituting~\eqref{eq:grad}, completing the square and adding the exact
measurement term yields the quadratic minorizer
\begin{equation}
  m_k(x\mid x_k)=-\frac{\|Ax-y\|_2^2}{2\beta_y^2}-\frac{\|x-\bar{x}_0(x_k)\|_2^2}{2\sigma_t^2}+c_k,
  \label{eq:minorizer}
\end{equation}
which satisfies $m_k\le C_t$ with equality in value and gradient at $x_k$. Its
exact maximizer solves
\begin{equation}
  \left(\frac{A^{\mathsf H}A}{\beta_y^2}+\frac{I}{\sigma_t^2}\right)x_{k+1}
  =\frac{A^{\mathsf H}y}{\beta_y^2}+\frac{\bar{x}_0(x_k)}{\sigma_t^2}.
  \label{eq:mstep}
\end{equation}
The system matrix is Hermitian positive definite for every $\sigma_t>0$ and is
applied inside a CG solver at a cost of one forward and one adjoint application
of $A$ per iteration. Only the prior term is minorized; the measurement term is
kept exactly, and no step size appears, since the weight $1/\sigma_t^2$
in~\eqref{eq:mstep} is the curvature constant of~\eqref{eq:curv} rather than a
free parameter. The expectation in~\eqref{eq:grad} is approximated by $S$
renoised samples.
%\begin{equation}
%  \tilde{x}_0(x_k)=\frac{1}{S}\sum_{s=1}^{S}\hat{x}_0\!\left(x_k+\sigma_t\varepsilon^{(s)}\right),
%  \quad \varepsilon^{(s)}\sim\mathcal{N}(0,I),
%  \label{eq:mc}
%\end{equation}
and~\eqref{eq:mstep} is solved with $L$ CG iterations. Algorithm~1 summarizes
the procedure.

\emph{Guarantees.} Under standard regularity conditions --- $G_t$ finite and
twice continuously differentiable with differentiation interchangeable with
expectation, the superlevel set $\{x:C_t(x)\ge C_t(x_0)\}$ compact, and
$\beta_y,\sigma_t>0$ --- the iteration~\eqref{eq:mstep} with the exact
expectation and an exact solve satisfies, for every $k$,
\begin{equation}
  C_t(x_{k+1})\;\ge\;C_t(x_k)+\frac{\|x_{k+1}-x_k\|_2^2}{2\sigma_t^2}.
  \label{eq:ascent}
\end{equation}
Hence $\{C_t(x_k)\}$ is nondecreasing and convergent, $\|x_{k+1}-x_k\|\to0$,
and every accumulation point satisfies
$A^{\mathsf H}(Ax^\star-y)/\beta_y^2+(x^\star-\bar{x}_0(x^\star))/\sigma_t^2=0$,
with no step-size condition and no Lipschitz constant required. As the
guaranteed increase grows with the square of the step and $C_t$ is bounded
above, the iterates settle at the fixed point rather than oscillate. The
argument needs $C_t$ to exist but never to be evaluated: $c_k$
in~\eqref{eq:minorizer} is fixed by $m_k(x_k)=C_t(x_k)$ and cancels, and
existence follows because~\eqref{eq:Gdef} is built from $p_{\sigma_t}$,
positive and smooth for every $\sigma_t>0$, whereas analyses defining the
regularizer through the denoiser need a conservative vector field. Replacing
$\bar{x}_0$ by the $S$-sample average~\eqref{eq:grad} from a learned denoiser of
bias bound $b_t$ and per-sample variance $v_t$ degrades~\eqref{eq:ascent} in
expectation only by $(v_t/S+b_t^2)/2\sigma_t^2$, the Monte Carlo term being
reducible by computation alone.

\emph{Annealing.} At large $\sigma_t$ the renoising perturbation dominates the
artifacts in the iterate, while at small $\sigma_t$ the gap contracts as
$O(\sigma_t^2)$, at the cost of increasingly local updates. We therefore use
$\sigma_{t_0}>\cdots>\sigma_{t_{T-1}}>0$, each level warm-started from the
preceding solution.

% =====================================================================
\vspace{-1em}
\section{Experiments}
\label{sec:exp}

%\subsection{Setup}
\emph{Data and prior.} We evaluate the algorithm on fastMRI
brain~\cite{zbontar2019fastmri}: axial validation subjects at $320\times320$
with 20 coils, 1D Cartesian undersampling with 12 ACS lines, the four corner
slices excluded, and 30 subjects per contrast (T1, T2, FLAIR) at $\times5$ and
$\times10$. All methods share a pretrained conditional EDM
prior~\cite{karras2022edm} trained on fastMRI brain data~\cite{tamirScore},
with the original noise schedule and timestep discretization.

\emph{Settings.} Hyperparameters are tuned per method (annealing steps,
noise range, schedule, measurement noise $\beta_y$); D-MOD uses $T=30$, $S=5$,
$K=1$, $L=10$. Baselines are the posterior samplers DPS~\cite{Chung2023DPS} and
DAPS~\cite{Zhang2025DAPS}, the optimization-based
DiffPIR~\cite{zhu2023diffpir} and DDS~\cite{chung2023dds}, and
SNORE~\cite{renaud2024snore} in its annealed proximal variant
SNORE-P~\cite{renaud2025proxconv}, the stronger of the two here (its analysis
assumes the denoiser is the gradient of a potential, which the shared EDM prior
does not provide). All methods use identical measurements and are timed on one
NVIDIA A100 GPU. Each finite-MMSE estimate averages $10$ reconstructions with
independent seeds, and D-MOD's AVG likewise; runtimes are per-slice, averaged
over contrasts and acceleration factors.
\vspace{-1em}
\subsection{Results}
Table~\ref{tab:fastmri} reports PSNR/SSIM for the three contrasts at $\times5$
and $\times10$ with per-slice runtime, and Fig.~\ref{fig:main}(a) plots
the quality--runtime trade-off for T1 at $\times5$. A single D-MOD reconstruction
reaches $36.07$\,dB on T1 at $\times5$ in $3.5$\,s, matching the best baseline
$10$-sample MMSE estimate (DiffPIR, $36.06$\,dB, $37.8$\,s) and beating the
rest at $11$--$73\times$ lower runtime. The pattern holds across
contrasts and both acceleration factors: single-sample D-MOD exceeds, in PSNR,
the MMSE estimates of DPS, DAPS and SNORE-P everywhere, and comes within
$0.31$\,dB of the strongest (DDS, DiffPIR) at an order of magnitude less
compute, trailing DiffPIR's MMSE SSIM by at most $0.009$; D-MOD's AVG attains
the best PSNR on T1 and FLAIR at both factors and on T2 at $\times5$. These MMSE
estimates are finite-sample approximations to the true ($S\to\infty$) MMSE,
which is why a single mode estimate can match or exceed them.

The reproducibility difference follows from how each method converges. D-MOD
performs MM at every noise level with the prior term averaged over $S$ draws,
so the prior variance falls as $O(1/S)$. The baselines do not converge to a
fixed point: DiffPIR and DDS take one stochastic denoiser step per level along
a fixed trajectory, and SNORE's constant-step updates fluctuate in a
seed-dependent neighborhood, so each run gives a different image. The SD maps of Fig.~\ref{fig:main}(b) make
this visible, with D-MOD~(AVG) showing the lowest pixel-wise variance, the
optimization-based methods next and the sampling-based methods highest,
independently of PSNR.

% =====================================================================
\vspace{-2em}
\section{Conclusion}
\label{sec:conclusion}

We presented D-MOD, a posterior mode-estimation framework for accelerated MRI
pairing a pretrained diffusion prior with data consistency through a
minorize--maximize iteration: a renoise--denoise--average step estimates the
gradient of a smoothed prior surrogate, and a quadratic CG update combines it
with the measurement model exactly. Annealing contracts the surrogate--target
gap as $O(\sigma_t^2)$, so the returned estimate is a stationary point of a
surrogate whose distance from the smoothed posterior is quantified at the
terminal noise level. On fastMRI brain, a single D-MOD reconstruction attains
the best single-sample PSNR in every reported setting and matches or beats
$10$-sample MMSE estimates at $11$--$73\times$ lower runtime, with the lowest
run-to-run variation of the methods compared. The mode is representative only
where the posterior is near-unimodal; at high acceleration, where distinct
modes are distinct plausible images, sampling is more informative.
\vspace{-1em}
\section{Acknowledgments}
\label{sec:ack}
This work is supported by NIH grants R01-AG067078, R01-EB031169, and
R01-EB019961.
\balance
\small
\bibliographystyle{IEEEbib}
\bibliography{dmod_refs}

\end{document}